\documentclass[conference]{IEEEtran}
\IEEEoverridecommandlockouts
\usepackage{cite}
\usepackage{amsmath,amssymb,amsfonts}
\usepackage{algorithmic}
\usepackage{algorithm}
\usepackage{graphicx}
\usepackage{textcomp}
\usepackage{bm}
\newtheorem{theorem}{\bf Theorem}
\newtheorem{myDef}{\bf Definition}
\usepackage{subfigure} 
\usepackage{utfsym}
\usepackage{caption} 

\usepackage{multirow}
\usepackage{siunitx}
\usepackage{booktabs}
\usepackage{xcolor}
\def\BibTeX{{\rm B\kern-.05em{\sc i\kern-.025em b}\kern-.08em
    T\kern-.1667em\lower.7ex\hbox{E}\kern-.125emX}}
\begin{document}

\title{Data-Driven Preference Sampling for Pareto Front Learning\\

}

\DeclareRobustCommand*{\IEEEauthorrefmark}[1]{%
    \raisebox{0pt}[0pt][0pt]{\textsuperscript{\footnotesize\ensuremath{#1}}}}
\author{
\IEEEauthorblockN{Rongguang Ye\IEEEauthorrefmark{1}, Lei Chen\IEEEauthorrefmark{2}, Weiduo Liao \IEEEauthorrefmark{1}, Jinyuan Zhang$^{*,}$\IEEEauthorrefmark{1}, \text{Hisao Ishibuchi}$^{*,}$\IEEEauthorrefmark{1}} \thanks{* Corresponding author.}

\IEEEauthorblockA{
\IEEEauthorrefmark{1} Guangdong Provincial Key Laboratory of Brain-inspired Intelligent Computation\\ Department of Computer Science and Engineering, Southern University of Science and Technology, Shenzhen 518055, China.}

\IEEEauthorblockA{
\IEEEauthorrefmark{2} School of Mathematics and Statistics, Guangdong University of Technology, Guangzhou 51000, China
}
\IEEEauthorblockA{\{yerg2023, liaowd\}@mail.sustech.edu.cn, chenlei3@gdut.edu.cn, \{zhangjy, hisao\}@sustech.edu.cn}
}

\maketitle
\begin{abstract}
Pareto front learning is a technique that introduces preference vectors in a neural network to approximate the Pareto front. Previous Pareto front learning methods have demonstrated high performance in approximating simple Pareto fronts. These methods often sample preference vectors from a fixed Dirichlet distribution. However, no fixed sampling distribution can be adapted to diverse Pareto fronts. Efficiently sampling preference vectors and accurately estimating the Pareto front is a challenge. To address this challenge, we propose a data-driven preference vector sampling framework for Pareto front learning. We utilize the posterior information of the objective functions to adjust the parameters of the sampling distribution flexibly. In this manner, the proposed method can sample preference vectors from the location of the Pareto front with a high probability. Moreover, we design the distribution of the preference vector as a mixture of Dirichlet distributions to improve the performance of the model in disconnected Pareto fronts. Extensive experiments validate the superiority of the proposed method compared with state-of-the-art algorithms.
\end{abstract}

\begin{IEEEkeywords}
Pareto Front Learning, Sampling, Dirichlet Distribution, Neural Network
\end{IEEEkeywords}

\section{Introduction}

In various real-world applications, multi-objective optimization (MOO) algorithms play a crucial role in addressing complex problems by optimizing multiple conflicting objectives simultaneously \cite{miettinen1999nonlinear}. For example, when dealing with engineering design problems, it is essential to incorporate factors such as efficient aerodynamic performance and structural robustness \cite{zhou2022multi}. MOO aims to obtain mutually non-dominated and are known as Pareto optimal solutions. Each Pareto optimal solution has an optimal trade-off, and all Pareto optimal solutions are known as the Pareto set. The image of the Pareto set in the objective space is known as the Pareto front.

In the past few decades, gradient-free MOO algorithms, such as multi-objective evolutionary algorithms \cite{deb2002fast,zhang2007moea} and particle swarm optimization algorithms \cite{eberhart1995particle}, have been successful in finding well-distributed solutions on the Pareto front. However, these algorithms may be less suitable for optimizing deep neural networks with a large number of parameters because they can be very time-consuming \cite{desideri2012multiple}. Many previous studies have pointed out that gradient-based MOO algorithms are more suitable for training large-scale neural networks \cite{dosovitskiy2020you,lin2020controllable,sener2018multi}. These algorithms train a single model to effectively handle multiple tasks simultaneously, called multi-task learning \cite{rich1997multitask}. In particular, multi-task learning linearly combines multiple objectives with a predefined preference vector to form a single objective. However, the use of a single predefined preference vector can only generate a corresponding Pareto optimal solution. Recent studies have explored using gradient-based methods that infinite Pareto optimal solutions and then obtain Pareto front, known as Pareto front learning \cite{navon2020learning}. These methods incorporate preference vectors into the feature space \cite{chen2022multi,ruchte2021scalable} or use a hypernetwork to learn different preferences \cite{lin2020controllable,navon2020learning}. The goal for each preference vector is to find a solution that is close to the Pareto front in the preference vector's direction.

\begin{figure}[t]
	\centering
\includegraphics[width=0.48\textwidth]{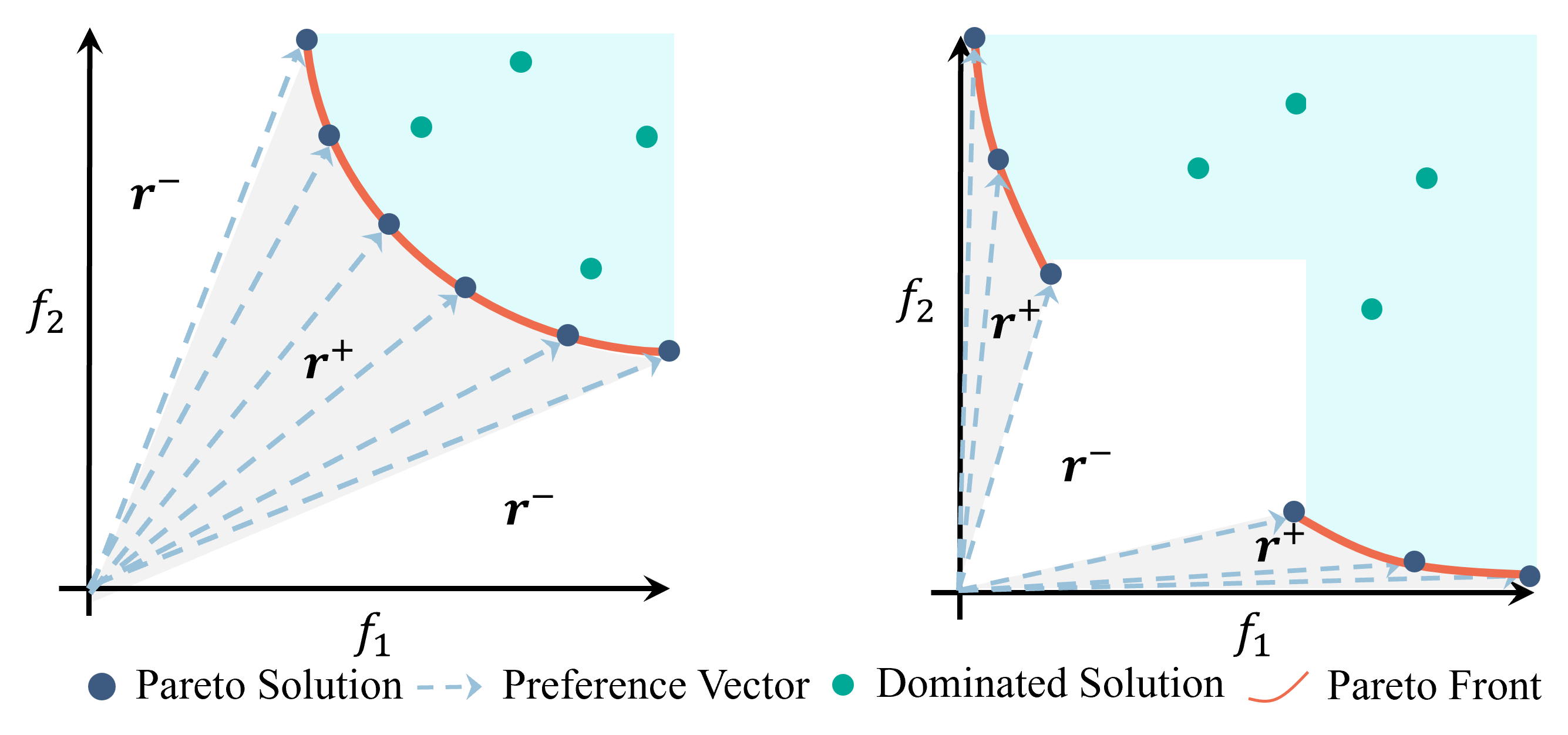}
	\caption{The blue region is the value range of the functions. We hope to find all the solutions of the Pareto front through different preference vectors.}
	\label{fig:motiv}
\end{figure}

Existing Pareto front learning methods have achieved significant performance in solving typical problems \cite{chen2022multi,liu2021profiling,ma2020efficient,ruchte2021scalable}. These methods assume that the Pareto front evenly covers the entire space. The preference vector are sampled from a predefined Dirichlet distribution to learn the Pareto front. However, the location of the Pareto front is uncertain, which will inevitably result in some redundant preference vectors. As shown in Fig. \ref{fig:motiv}, In a multi-objective minimization problem, $r^-$ represents invalid preference vector sampling regions, while $r^+$ represents an effective preference vector sampling region. This is because the preference vector in the region $r^+$ has corresponding Pareto optimal solutions. The Pareto front in the right figure is disconnected. In this case, we should sample the preference vector at $r^+$ rather than the disconnected region ($r^-$) of the Pareto front. However, the current Pareto front learning methods often use uniform preference vector sampling, which will reduce the performance of the neural network for Pareto front estimation.

To improve the accuracy of estimating different Pareto fronts, this paper proposes a Data-Driven Preference Vector Sampling (DDPS-MCMC). Specifically, the objective values of solutions generated during the training process are used as posterior information. Based on this posterior information, we use Markov chain Monte Carlo to adjust the parameters of the sampling distribution. The use of the posterior information allows us to sample the preference vector on the Pareto front with a high probability. Moreover, we design the sampling distribution as a mixture of Dirichlet distributions, it can be applied not only to the continuous Pareto front but also to the disconnected Pareto front. The main contributions of this paper can be summarized as follows:
\begin{itemize}
    \item  We propose a Data-Driven Preference Vector Sampling (DDPS-MCMC) for Pareto front learning. This method can continuously learn the parameters of the sampling distribution. In this way, the preference vectors can be adaptively sampled based on the location of the Pareto front.
    \item We propose a novel sampling distribution formulated as a mixture of Dirichlet distributions, which can flexibly learn different types of Pareto fronts.
    \item We conduct extensive experiments and comparisons on different test problems to demonstrate the effectiveness of the proposed method in handling various types of Pareto fronts.

\end{itemize}

\section{Preliminary \label{S2}}

A multi-objective optimization problem (MOP) with $m$ objectives can be defined as 
 follows:

\begin{equation}
    \min  \mathcal{L}(\bm{x})=\left({\mathcal{L}_{1}(\bm{x})}, {\mathcal{L}_{2}(\bm{x})}, \ldots, {\mathcal{L}_{m}(\bm{x})}\right),
\end{equation}
where $\bm{x} \in \mathbb{R}^{d}$, $\mathcal{L}(\bm{x})$ is a $m$-dimensional vector that characterizes the quality of $\bm{x}$ in $m$ objectives. Some definitions of multi-objective optimization are follows:

 \begin{myDef} \textbf{(Pareto Dominance).}
  A solution $\bm{x_{a}}$ is said to dominate another solution $\bm{x_{b}}$, denoted as $\bm{x_{a}}\prec \bm{x_{b}}$, if $\mathcal{L}_{i}(\bm{x_{a}}) \leq  \mathcal{L}_{i}(\bm{x_{b}}), \forall i \in \{1,...,m\}$ and $\exists j \in \{1,...,m\}$ such that $\mathcal{L}_{j}(\bm{x_{a}}) < \mathcal{L}_{j}(\bm{x_{b}})$.
\end{myDef}
 
 \begin{myDef} \textbf{(Pareto optimal solution).}
 Solution $\bm{x_{a}}$ is called a Pareto optimal solution if there does not exist a solution $\bm{x_{b}}$ such that $\bm{x_{b}} \prec \bm{x_{a}}$.
 \end{myDef}

\begin{myDef} \textbf{(Pareto Set / Front).}
The set of all Pareto optimal solutions is called the Pareto set, denoted by $\mathcal{M}_{ps}$. The image of the Pareto set in the objective space is called the Pareto front  $\mathcal{P}= \{\mathcal{L}(\bm{x}), \bm{x} \in$ $\mathcal{M}_{ps}$ $\}$. 
 \end{myDef}
 
 A Pareto optimal solution can reflect a user's preference information for multiple objectives. The generation of a Pareto optimal solution requires incorporating a preference vector $\boldsymbol{r}$ into the model, where the preference vector $\boldsymbol{r}=[r_{1}, r_{2},..., r_{m}] \in\mathbb{R}^{m}$. Specifically, given a training distribution $\mathbb{P}_{\bm{x},\bm{y}}$ of pairs $(\bm{x},\bm{y})$ where $\bm{x} \in \mathbb{X}\subset \mathbb{R}^{d_\mathbb{X}}$, $\bm{y}\in \mathbb{Y}\subset \mathbb{R}^{d_\mathbb{Y}}$, and a loss function $\mathcal{L}_{i} : \mathbb{Y}\times \mathbb{Y}\to \mathbb{R}$, we aim to learn a model $F: \mathbb{X}\to \mathbb{Y}$ with parameters $\bm{\theta}$ that minimizes the expected loss $\mathbb{E}_{(\bm{x},\bm{y})\sim \mathbb{P}_{\bm{x},\bm{y}}} \mathcal{L} (y, F(\bm{x};\bm{\theta}), \boldsymbol{r})$ over the dataset to obtain all Pareto optimal solutions.\footnote{$d_\mathbb{X}$: Dimensions of $\mathbb{X}$ space, $d_\mathbb{Y}$: Dimensions of $\mathbb{Y}$ space}

\begin{figure*}[t]
    \centering
    \includegraphics[height=1.35in]{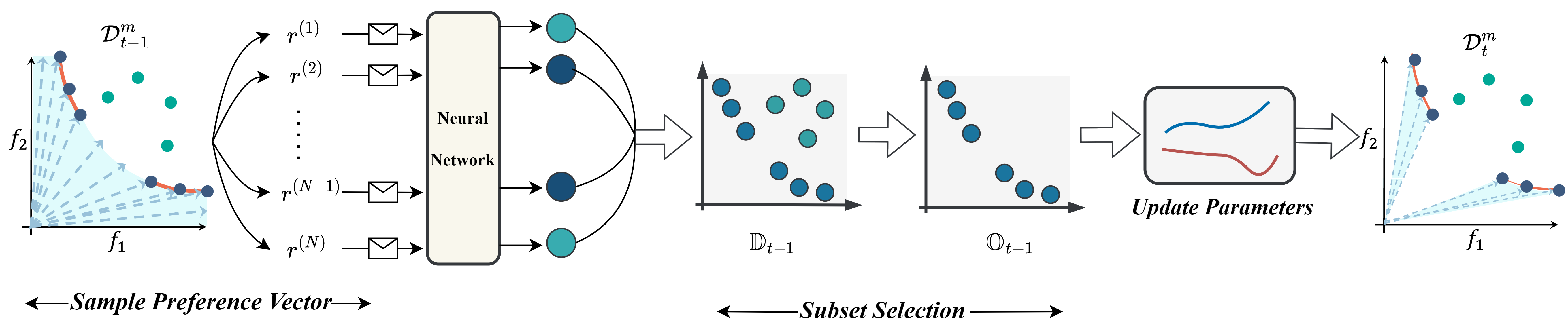}
    \caption{Data-driven Preference Sampling (DDPS) Flow Chart. In the ($t-1$)-th epoch of training, the preference vectors are sampled from the mixture of Dirichlet distributions $\mathcal{D}^{m}_{{t-1}}$ to form the set of loss values $\mathbb{D}_{t-1}$, which contains $N$ samples. After subset selection, the best $\lfloor \gamma pN \rfloor$ preference vectors are selected. The mixture of Dirichlet distributions is finally updated based on the posterior information.}
    \label{fig:frame}
\end{figure*}

Let us give two examples to clearly understand the Pareto front learning task in different scenarios.
\begin{itemize}
    \item  We have a large amount of image data when applying Pareto front learning in multi-task image classification. In this case, one possible method is to embed the preference vector into the feature map. The scenario at this time is how to learn the representation of the neural network based on different preference vectors.
    \item We do not have training data when applying Pareto front learning to multi-objective optimization benchmark problems. In this case, we can design the output of the neural network as decision variables. The corresponding loss function depends on the calculation of objective functions. The scenario at this time is how to obtain different decision values based on different preference vectors.
\end{itemize}
The Pareto front learning method can obtain optimal trade-off solutions by sampling different preference vectors in a neural network framework. Preference vectors are sampled from a Dirichlet distribution ($\mathcal{D}$) in each batch to incorporate variability in preferences into the model. Preference vectors are used as weights and applied to the loss functions of multiple objectives. The most direct optimization approach is linear scalarization: 

\begin{equation}
    \boldsymbol{\theta}_{\boldsymbol{r}}^{*}=\underset{\boldsymbol{\theta}}{\arg \min }  \ \mathbb{E}_{\boldsymbol{r} \sim \mathcal{D}}\mathbb{E}_{\boldsymbol{x}, \boldsymbol{y} \sim \mathbb{P}_{\boldsymbol{x}, \boldsymbol{y}}} \boldsymbol{r} \mathcal{L}(\boldsymbol{y}, F(\boldsymbol{x}, \bm{\theta}, \bm{r})).
\end{equation}



\section{Data-Driven Preference Sampling \label{S3}}
Preference vectors are often sampled from the Dirichlet distribution $\mathcal{D}(\bm{\alpha})$ in previous studies \cite{hoang2023improving,navon2020learning,ruchte2021scalable}, where $\bm{\alpha}$ is predefined. We use the data generated by the training process as posterior information. To improve the effectiveness of utilizing preference vectors, we propose a novel framework for updating sampling distributions based on posterior information. Our core idea is to adaptively adjust the parameters of the sampling distribution through the posterior information of the previous epochs. As shown in Fig. \ref{fig:frame}, each preference vector in the ($t-1$)-th epoch will generate a corresponding solution when they are inputted into the neural network. Afterward, useful preference vectors are selected to update the parameters of the sampling distribution at the $t$-th epoch (details in Section \ref{3.2}). The optimization goal of the neural network in the $(t-1)$-th epoch is as follows:
\begin{equation} \label{eq3}
    \underset{\boldsymbol{\theta}}{\min } \ \mathbb{E}_{\boldsymbol{r} \sim \mathcal{D}(\bm{\alpha}^{(t-1)})}\mathbb{E}_{\boldsymbol{x}, \boldsymbol{y} \sim \mathbb{P}_{\boldsymbol{x}, \boldsymbol{y}}} \mathcal{L}(\boldsymbol{y}, F(\boldsymbol{x}, \bm{\theta}), \boldsymbol{r}).
\end{equation}

In the following subsections, we will discuss how to design and collect data to learn the sampling distribution in Equation (\ref{eq3}).

\subsection{Sampling Distribution Designing}
The probability density function of the Dirichlet distribution is defined as follows:
\begin{equation}
\mathcal{D}\left(X \mid \boldsymbol{\alpha}\right)=\dfrac{1}{\mathrm{B}\left(\boldsymbol{\alpha}\right)}\prod\limits_{i=1}^{m}X_{i}^{\alpha_{i}-1},
\end{equation}

\begin{theorem} \label{theorem}
  Let $X$ be a random vector that follows a Dirichlet distribution $\mathcal{D}$ with parameters $\bm{\alpha} = (\alpha_{1}, \alpha_{2}, ..., \alpha_{m})$. Then, the expected value $\mathbb{E}(x_{k})$ and variance $\mathbb{V}(x_{k})$ satisfy   \\ \centerline{$\mathbb{E}\left[x_{k}\right]=\dfrac{\alpha_k}{\sum_i\alpha_i},\mathbb{V}\left[x_{k}\right]=\dfrac{\alpha_k\left(\sum_i\alpha_i-\alpha_k\right)}{\sum_i\alpha_i^2\left(\sum_i\alpha_i+1\right)}$,}
\end{theorem}
where $B(\bm{\alpha})$ is multivariate Beta function, $||X||_{1}$ = 1 and $X > 0$. 
According to the properties of the expectation and variance of the Dirichlet distribution in Theorem \ref{theorem}, we can assume a symmetric Dirichlet prior of the form $\alpha_k = \varepsilon /m$, where $\varepsilon \in \mathbb{R}^{+}$. In this case, the expected value of the Dirichlet distribution is $\mathbb{E}\ [x_{k}] = 1/m$, and the variance is $\mathbb{V}\ [x_{k}] = \frac{m-1}{m^2(\varepsilon+1)}$. The larger the $\varepsilon$, the smaller the variance, and the more concentrated the distribution. Different Pareto fronts have a corresponding suitable $\bm{\alpha}$.
However, the unimodal Dirichlet distribution is not suitable for solving disconnected Pareto fronts (Fig. \ref{fig:motiv}). To extend our method to a more complex Pareto front, we define the mixture of Dirichlet distributions $\mathcal{D}^{m}$:

\begin{equation}
    \mathcal{D}^{m}\left(X \mid \boldsymbol{\alpha}, \bm{\omega^{(j)}} \right) = \sum_{i=1}^\kappa w^{(j)}_{i} \mathcal{D}_i(X\mid \bm{\alpha}),
\end{equation}
where $\bm{\omega^{(j)}}$ is the weight of the $j$-th epoch, $\sum\nolimits_{i=1}^\kappa \omega_i=1$ and $\kappa$ is the number of components in the mixture of Dirichlet distributions. The mixture of Dirichlet distribution is a multimodal distribution rather than just an unimodal distribution so that it can adapt flexibly to complex Pareto fronts. In the next subsection, we will introduce how to collect posterior information to learn $\omega^{(j)}$ and $\bm{\alpha}$.

\subsection{Collecting Posterior Information
 \label{3.0}}
As shown in Fig. \ref{fig:frame}, during the training process, the preference vector $\boldsymbol{r}^{(i)}$ ($i=1,2,...,N$) is sampled from the mixture of Dirichlet distributions $\mathcal{D}^{m}_{t-1}$. Each $\boldsymbol{r}^{(i)}$ can obtain a loss function vector by the forward propagation of the neural network, i.e., $\mathcal{L}(\bm{\theta}, \boldsymbol{r}^{(i)})=(\mathcal{L}_{1}(\bm{\theta}, \boldsymbol{r}^{(i)}),...,\mathcal{L}_{m}(\bm{\theta}, \boldsymbol{r}^{(i)}))$. Let us denote the loss matrix collected from the ($t-1$)-th epoch as $\mathbb{D}_{t-1}=\left[\mathcal{L}_{n1},\ldots,\mathcal{L}_{nm}\right]_{n=1}^N$. We need to find a preference vector that can generate a Pareto set to provide accurate information for updating the sampling distribution of the next epoch. We select useful preference vectors from $\mathbb{D}_{t-1}$ to update $\alpha^{(t-1)}$ and $\omega^{(t-1)}$, since not all data in $\mathbb{D}_{t-1}$ is useful. Promising solutions can be selected according to \textit{Non-Dominated Sorting (NDS)} and \textit{Crowding Distance (CD)} in multi-objective optimization algorithm NSGA-II \cite{deb2002fast}. The calculation of dominance ranking is needed in \textit{NDS}, which is defined as follows:

\begin{myDef} Let $\mathbb{D}=\{\mathbf{a}_1, \ldots, \mathbf{a}_N\}$ be a set of solutions in a multi-objective optimization problem. The dominance rank $\mathbb{I}(\mathbf{a}_i)$ of solution $\mathbf{a}_i$ is defined as:

\[
\mathbb{I}(\mathbf{a}_i) = \sum_{\mathbf{a}_j \in \mathbb{D}, \mathbf{a}_j \neq \mathbf{a}_i} \mathbf{a}_j \prec \mathbf{a}_i.
\]
\end{myDef}

In simple terms, \textit{NDS} ranks the dominance ranking (the smaller the $\mathbb{I}$, the stronger the solution) of solutions in $\mathbb{D}_{t-1}$, while \textit{CD} calculates the distances between the solutions. At the same rank of dominance, the \textit{CD} value of a solution needs to be compared (the larger the \textit{CD}, the more information the data point has). We select $\lfloor\gamma pN\rfloor$ useful preference vectors from the dataset $\mathbb{D}_{t-1}$ based on \textit{NDS} and \textit{CD} (donated as \textit{NDS-CD}):  
\begin{equation} \label{nds}
\begin{aligned}
    & \mathbb{O}_{t-1} \gets NDS\text{-}CD(\mathbb{D}_{t-1}, \lfloor \gamma pN \rfloor),
\end{aligned}
\end{equation}
where $\gamma\in(0,\ 1)$ is used to select an upper bound on the amount of data ($\mathbb{O}_{t-1}$), and $p$ represents the current epoch number of the model training.

After obtaining the observed samples $\mathbb{O}_{t-1}$, we maximize the posterior probability to learn $\bm{\alpha^{(t)}}$ and $\bm{\omega^{(t)}}$. Then, the mixture of Dirichlet distributions $\mathcal{D}^{m}(\bm{\alpha^{(t-1)}})$ is updated for the next epoch. In general, the optimization problem we consider in the $t$-th epoch is as follows:

\begin{equation}
    \begin{aligned}
&\underset{\bm{\theta}}{\operatorname{min}} \ \mathbb{E}_{\boldsymbol{r}\sim\mathcal{D}^{m}(\bm{\alpha^{(t)}}, \bm{\omega^{(t)}})}\mathbb{E}_{\bm{x}, \bm{y}\sim\mathbb{P}_{\bm{x},\bm{y}}}\mathcal{L}(\bm{y}, F(\bm{x},\bm{\theta}),\boldsymbol{r}) \\
& \quad \mathrm{ s.t. } \ \bm{\alpha^{(t)}}, \bm{\omega^{(t)}}=\underset{\bm{\alpha}, \bm{\omega}}{\operatorname{argmax}} \ \mathcal{D}^{m}(\bm{\alpha}, \bm{\omega} \mid \mathbb{O}_{t-1}).
\end{aligned}
\end{equation}
How to solve this optimization problem is a main issue. In the next subsection, we will utilize Bayesian inference to learn $\bm{\alpha^{(t)}}$ and $\bm{\omega^{(t)}}$.

\begin{figure}[t]
    \centering
    \includegraphics[height=1.1in]{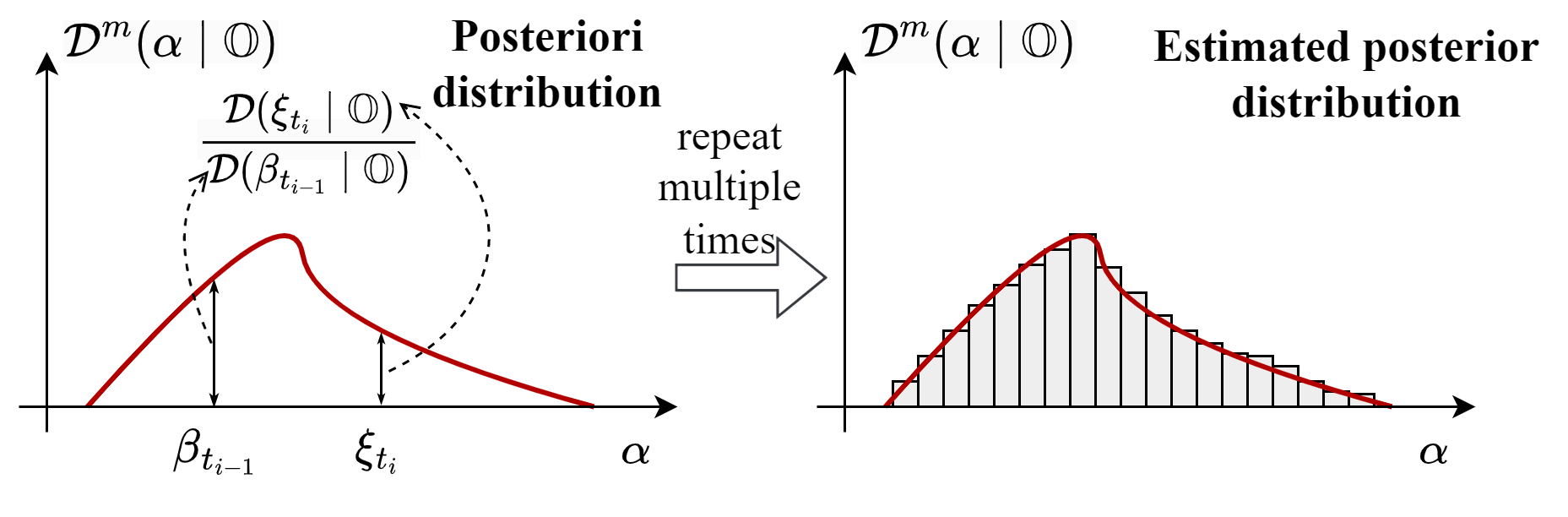}
    \caption{Sampling process of MCMC.}
    \label{fig:mcmc}
\end{figure}

\subsection{Sampling Distribution Learning\label{3.2}}

In this subsection, we present the Markov Chain Monte Carlo (MCMC) method to learn the mixture of Dirichlet distributions. The most popular method in MCMC is the Metropolis-Hastings algorithm \cite{chib1995understanding}. As mentioned in previous subsection, $\mathbb{O}_{t-1}$ is obtained in the ($t-1$)-th epoch. MCMC is a technique for repeatedly sampling parameters to be estimated from a given prior distribution. The proposal of the $i$-th sampling is represented as ${\bm{\xi}}_{t_{i}}=(e^{\bm{\widehat{\alpha}^{(t_{i})}}}, \bm{\widehat{\omega}^{(t_{i})})}$. Among them,  $\bm{\widehat{\alpha}^{(t_{i})}}$ and $\bm{\widehat{\omega}^{(t_{i})}}$ are sampled from $\mathcal{N}(\bm{\widehat{\alpha}^{(t_{i})}} \mid \bm{\mu}, \bm{\sigma})$ and Dirichlet distribution $\mathcal{D}(\bm{\alpha}=\textbf{1})$, respectively. The posterior Dirichlet distribution of the current $t$-th epoch is denoted as $\mathcal{D}^{m}(\bm{{\xi}_{t_{i}}}\mid \mathbb{O}_{t-1})$. Then, we calculate the acceptance probability of the $i$-th proposal:

\begin{equation}
    p_t=\min(\frac{\mathcal{D}^{m}(\bm{\xi_{t_i}} \mid \mathbb{O}_{t-1})}{\mathcal{D}^{m}(\bm{\beta_{t_{i-1}}} \mid \mathbb{O}_{t-1})}, 1),
\end{equation}
where $\bm{\beta_{t_{i-1}}}$ is the previous accepted proposal. According to the Bayes theorem, the acceptance probability $p_{t}$ of the final proposal $\bm{\xi_{t_i}}$ is:
\begin{equation}
    p_t=\min(\frac{\mathcal{D}(\mathbb{O}_{t-1} \mid \bm{\xi_{t_i})}}{\mathcal{D}(\mathbb{O}_{t-1}\mid \bm{\beta_{t_{i-1}})}}\dfrac{\mathcal{N}(\bm{\xi_{t_i}})\mathcal{D}(\bm{\xi_{t_i})}}{\mathcal{N}(\bm{\beta_{t_{i-1}}})\mathcal{D}(\bm{\beta_{t_{i-1}})}},1).
\end{equation}

All the terms can be calculated by plugging them into their corresponding probability density function. Finally, a random variable $k_t \in \left(0, 1 \right)$ is drawn from a uniform distribution, 
and if $p_t>k_{t}$, then $\bm{\xi_{t_{i}}}$ is accepted ($\bm{\beta_{{t}_{i}}}=\bm{\xi_{{t}_{i}}}$). As shown in Fig. 3, the higher the posterior probability of a proposal $\bm{\xi_{{t}_{i}}}$, the higher the probability of acceptance. When the sampling is repeated multiple times, the entire posterior distribution can be estimated. The Markov chain is not stable in the early stage. The samples with large discrepancies are called burn-in samples. Therefore, we extract the expectation of the Markov chain from the latter half as the parameter of the Dirichlet distribution for the $t$-th epoch. That is $\bm{\alpha^{(t)}} = \mathbb{E}(e^{\bm{\widehat{\alpha}^{(t)}_{i}}}); \bm{\omega^{(t)}} = \mathbb{E}(\widehat{\bm{\omega}}^{(t)}_{i})$, where $i = \frac{S}{2},\ldots,S$, and $S$ is the number of Monte Carlo samples.

Our proposed method maximizes the posterior mixture of Dirichlet distributions based on the data observed in each epoch. In this manner, we sample promising preference vectors with a higher probability and other regions with a lower probability. The pseudocode of our framework is shown in Algorithm \ref{ddps}.

\begin{algorithm}[t]
\caption{Data-Driven Preference Sampling (DDPS)}\label{ddps}
\textbf{Input}: $\bm{\alpha^{(t-1)}}$, $\bm{\omega^{(t-1)}}$ and current epoch $t$\\
\textbf{Parameter}: $\gamma \in (0, 1)$, $\bm{\mu}$, $\bm{\sigma}$ and $S$.\\
\textbf{Output}: $\bm{\alpha^{(t)}}, \bm{\omega^{(t)}}$
\begin{algorithmic}[1] 
\STATE $\bm{\beta_{t_{0}}} = (\bm{\alpha^{(t-1)}}, \bm{\omega^{(t-1)}})$
\STATE $\mathbb{D}_{t-1}=\left[\mathcal{L}_{n1},\ldots,\mathcal{L}_{nm}\right]_{n=1}^N$ 
\STATE $\mathbb{D}_{t-1} =  \mathbb{D}_{t-1} / $ $sum(\mathbb{D}_{t-1},axis=1)$  $\rhd$ Normalization
\STATE $\mathbb{O}_{t-1} \gets NDS\text{-}CD(\mathbb{D}_{t-1}, \lfloor \gamma pN \rfloor)$
\FOR{$i=$1 \textbf{$to$} S} 

    \STATE $\bm{\widehat{\alpha}^{(t_{i})}} \sim \mathcal{N}\left(\bm{\mu},\bm{\sigma}\right)$, \quad $\bm{\widehat{\omega}^{(t_{i})}} \sim \mathcal{D}(\bm{\alpha}=\textbf{1})$
    \STATE $\xi_{t_i}=(e^{\bm{\widehat{\alpha}^{(t_{i})}}}, \bm{\widehat{\omega}^{(t_{i})}})$
    \STATE  $p_i=\min(\frac{\mathcal{D}^{m}(\bm{\xi_{t_i}} \mid \mathbb{O}_{t-1})}{\mathcal{D}^{m}(\bm{\beta_{t_{i-1}}} \mid \mathbb{O}_{t-1})}, 1)$
    \STATE $k_i \sim \mathcal{U}(0, 1)$ 

    \IF{$k_i\le p_i$}
        \STATE  $\bm{\beta_{t_i}}=\bm{\xi_{t_i}}$
    \ELSE
    \STATE  $\bm{\beta_{t_i}}= \bm{\beta_{t_{i-1}}}$ 
\ENDIF
\ENDFOR
\\
$\bm{\alpha^{(t)}}, \bm{\omega^{(t)}} =\mathbb{E}\left(\bm{\beta_{t_i}}\right),\ i=\frac{S}{2},\ldots,S$
\STATE \textbf{return} $\bm{\alpha^{(t)}}, \bm{\omega^{(t)}}$
\end{algorithmic}
\end{algorithm}

\section{Experiments \label{S4}}

\begin{table*}[t]
\centering
\caption{Comparison of our method with state-of-the-art methods in terms of HV and IGD value on synthetic problems.}
\label{s}
\renewcommand{\arraystretch}{1.2}
\setlength{\tabcolsep}{1.8pt}
\begin{tabular}{
    c
    S[table-format=1.4]
    S[table-format=1.4]
    S[table-format=1.4]
    S[table-format=1.4]
    S[table-format=1.4]
    S[table-format=1.4]
    S[table-format=1.4]
    S[table-format=1.4]
    S[table-format=1.4]
    S[table-format=1.4]
    }
\toprule
\multirow{2}{*}{Problem} & \multicolumn{5}{c}{HV $\uparrow$} & \multicolumn{5}{c}{IGD $\downarrow$} \\
\cmidrule(lr){2-6} \cmidrule(lr){7-11}
& {PHN-LS} & {COSMOS} & {NORM} & {PHN-HVI} & {DDPS-MCMC} & {PHN-LS} & {COSMOS} & {NORM} & {PHN-HVI} & {DDPS-MCMC} \\
\midrule
LZLZK & 2.9791 & \textbf{3.349} & 3.2913 & 3.1811 & 3.3199 & 0.0871 & \textbf{0.0067} & 0.0132 & 0.0349 & 0.0176 \\
ZDT3 & 0.8158 & 1.6403 & 2.8065 & 2.4929 & \textbf{2.8365} & 0.0245 & 0.4232 & 0.0169 & 0.1225 & \textbf{0.0034} \\
DTLZ4 & 2.2012 & 2.2434 & 2.4815 & 2.6512 & \textbf{2.7745} & 0.3945 & 0.1982 & 0.1375 & 0.0775 & \textbf{0.0407} \\
DTLZ5 & \textbf{6.1020} & 5.6961 & 5.6794 & 5.7863 & 5.6506 & \textbf{0.0035} & 0.0164 & 0.0094 & 0.0400 & 0.0102 \\
DTLZ7 & 5.2337 & 1.9352 & 5.2974 & 6.1504 & \textbf{6.2968} & 0.3444 & 0.9012 & 0.2725 & 0.0438 & \textbf{0.0286} \\
\midrule
Average Rank & 4.0 & 3.4 & 3.0 & 2.6 & \textbf{2.0} & 3.6 & 4.7 & 2.4 & 3.4 & \textbf{1.8} \\
\bottomrule
\end{tabular}
\end{table*}

\begin{figure*}[t]
    \centering
    \includegraphics[width=0.98\textwidth]{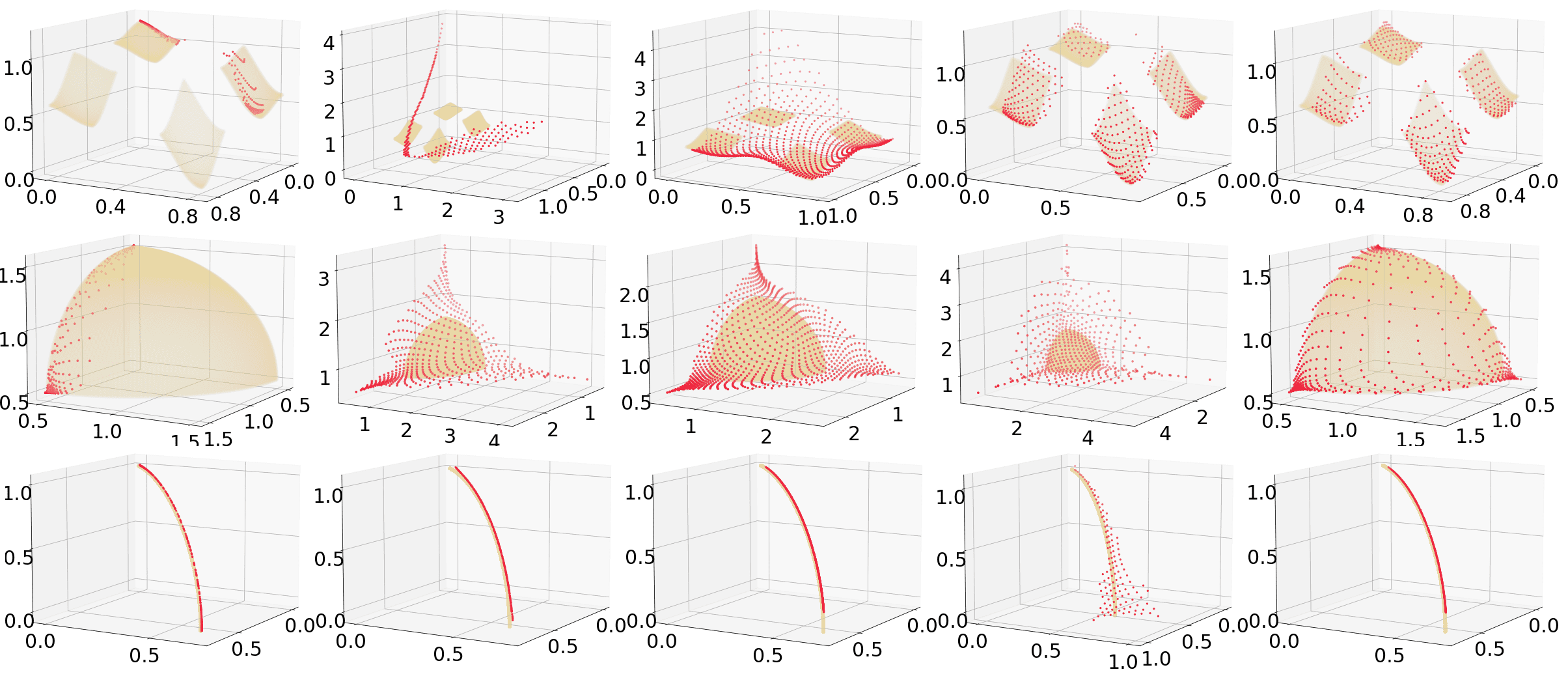}
    \caption{Plot for three-objective problems. The first row is DTLZ7, the second row is DTLZ4, and the third row is DTLZ5. The yellow surfaces are the Pareto fronts, while the orange points are non-dominated solutions.}
    \label{fig:problem2}
\end{figure*}

\subsection{Experimental Settings\label{sec.31}}

\subsubsection{Dataset}
The proposed algorithm is tested on five synthetic test problems, including two two-objective problems, i.e., ZDT3 \cite{zitzler2000comparison}, LZLZK \cite{lin2019pareto}, and three three-objective problems, i.e., DTLZ4, DTLZ5, and DTLZ7 \cite{deb2005scalable}. The Pareto fronts of these problems are shown in Fig. \ref{fig:problem2} and Fig. \ref{fig:problem1}.
The Pareto front of DTLZ4 and LZLZK is continuous, while the Pareto front of DTLZ7 and ZDT3 is disconnected. The Pareto front of DTLZ5 is degenerated. Three image classification datasets (i.e., Multi-MNIST, Multi-Fashion, and Multi-Fashion+MNIST \cite{xiao2017fashion}) are also used to evaluate the proposed algorithm. Each of these datasets poses a multi-task learning challenge with two objectives: categorizing the top-left (task left) and bottom-right (task right) items. Each test set contains 20,000 images, and the training set consists of 120,000 images, with 10\% allocated for validation.

\subsubsection{Evaluation Metric}
Hypervolume (HV) \cite{miettinen1999nonlinear} and Inverted Generational Distance (IGD) \cite{coello2004study} are used to evaluate the quality of the learned Pareto front. IGD measures the difference between the true Pareto front and the approximated Pareto front. The smaller the IGD, the better the approximated performance of the learned Pareto front. The higher the HV, the better the Pareto front. For HV caculation only a reference point needs to be provided. Since the Pareto front of each multi-task image classification problem is unknown, we only use HV to evaluate the performance of the approximated Pareto front for each problem. The reference point for HV calculation is (2, 2).

\subsubsection{Baselines}
Five PFL methods are used as baseline algorithms including PHN-HVI \cite{hoang2023improving}, PHN-LS \cite{navon2020learning}, PHN-EPO, COSMOS \cite{ruchte2021scalable} and NORM \cite{mausser2006normalization} based on the YOTO \cite{dosovitskiy2020you} methods.

\subsubsection{Implementation Details}
The LeNet architecture \cite{sener2018multi} is used for all methods. The parameters $\bm{\mu}$ and $\bm{\sigma}$ of the normal distribution are set to $\bm{0}$ and $\bm{2}$, respectively. The number of proposal samples randomly drawn in MCMC is set to 10000, and the early stop strategy is used during training. The maximum number of epochs is 1000 for synthesized problems and 60 for image classification problems. All methods are evaluated using the same distributed preference vectors. Moreover, YOTO \cite{dosovitskiy2020you} is a method that can incorporate preference vectors as inputs for training models. In this work, we utilize YOTO to merge preference vectors and employ a penalty-based boundary loss function \cite{li2019adjustment}. All experiments are conducted in PyTorch \cite{paszke2019pytorch}, and an NVIDIA GeForce RTX 3090 GPU with 24G of memory is used for training.

\begin{table*}[t]
\centering
\renewcommand{\arraystretch}{1.2}
\setlength{\abovecaptionskip}{0pt}%
\caption{Comparison of DDPS-MCMC with advanced methods in terms of hypervolume, runtime, and number of parameters. \label{visionft}}
\begin{tabular}{cc|cccccc}
\hline
\multirow{2}{*}{Method} & \multirow{2}{*}{\#Params.} & \multicolumn{2}{c}{Multi-Fashion} & \multicolumn{2}{c}{Multi-MNIST} & \multicolumn{2}{c}{Fash. + MNIST} \\ &           & HV              & Times (Sec.)     & HV              & Times (Sec.)   & HV               & Times (Sec.)    \\ \hline
COSMOS                  & 4.3K                       & 2.32            & \textbf{194}             & 2.96            & \textbf{187}           & 2.86             & \textbf{328}   \\
PHN-LS                  & 3243K                      & 2.13            & 401             & 2.84            & 388           & 2.77             & 527            \\
PHN-EPO                 & 3243K                      & 2.23            & 1385            & 2.84            & 864           & 2.74             & 882            \\ 
PHN-HVI                & 3243K                       & 2.27             & 3801    & 2.94            & 3743           & 2.81    & 4016            \\
DDPS-MCMC               & 4.4K                       & \textbf{2.40}    & 214             & \textbf{3.06}   & 217           & \textbf{2.96}             & 419            \\ \hline
\end{tabular}
\end{table*}

\begin{figure*}[ht]
    \centering
    \includegraphics[width=0.98\linewidth]{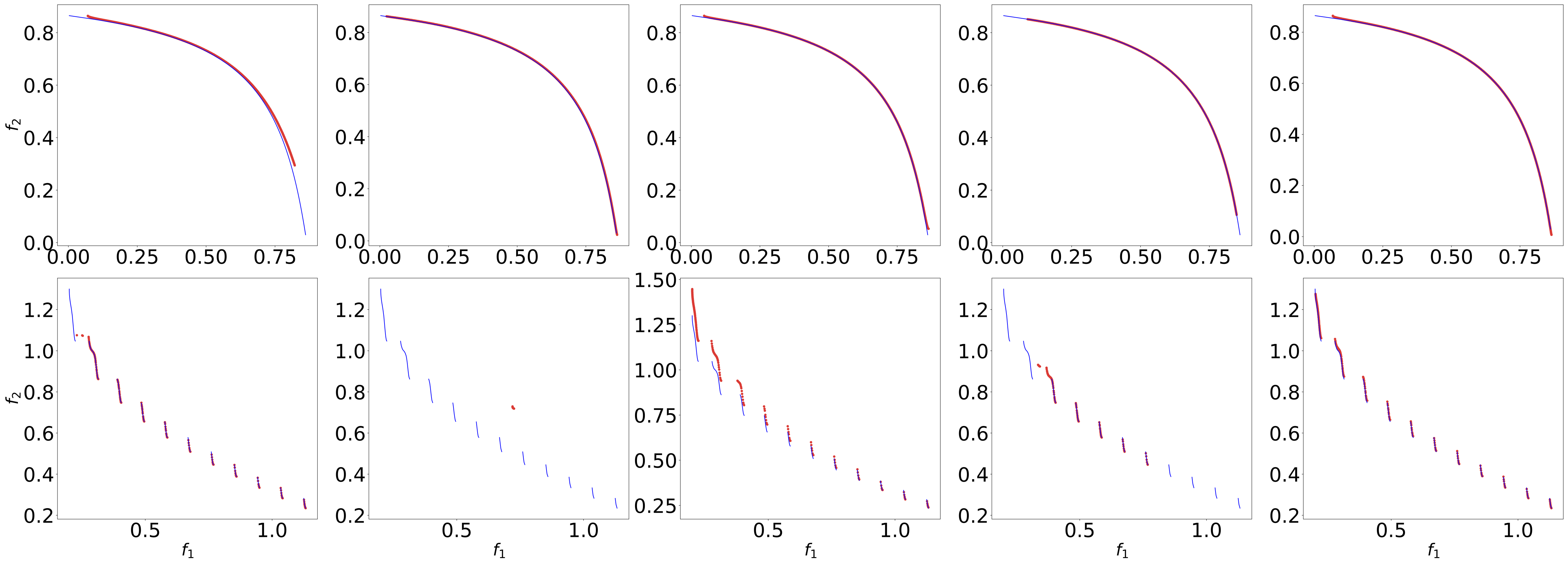}
    \caption{Plot of two-objective problems. The first row is LZLZK, and the second row is ZDT3. The blue curves are the true Pareto front and the red points indicate the set of non-dominated solutions.}
    \label{fig:problem1}
\end{figure*}

\begin{figure*}[t]
    \centering
    \includegraphics[width=0.9\linewidth]{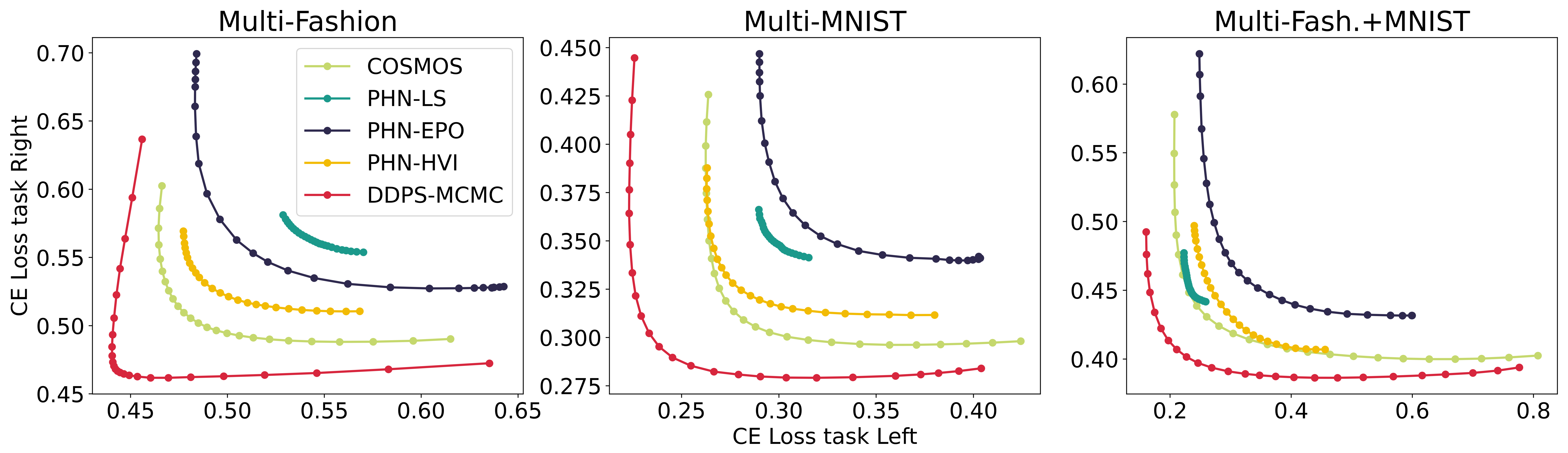}
    \caption{Comparison of the Pareto front generated by the proposed method and state-of-the-art methods.}
    \label{fig:comp}
\end{figure*}

\subsection{Experimental Results}
\subsubsection{Synthetic Problems \label{sec32}}
As shown in Table \ref{s}, the proposed method achieves the best performance with average ranks of 1.8 for IGD value and 2 for HV value across the five test problems. The Pareto front obtained by each algorithm for each of the three objective problems is shown in Fig. \ref{fig:problem2}. 
The Pareto front of DTLZ7 is divided into four areas, indicating that the effective sampling space of preference vectors is also divided into four regions. The results show that DDPS-MCMC has the best performance, followed by PHN-HVI. Since NORM cannot handle the problem with disconnected Pareto fronts, some solutions are scattered far from the Pareto front. The results of the proposed method accurately cover the solutions on the Pareto front for DTLZ4. Most of these algorithms have good convergence performance in DTLZ5, but some solutions of PHN-HVI exhibit poor convergence ability. 

The obtained Pareto fronts of two-objective problems are shown in Fig. \ref{fig:problem1}. For the continuous test problem (LZLZK), most algorithms can accurately approximate the Pareto front. While PHN-LS and PHN-HVI perform slightly worse. The performance of the five baselines is estimated to be worse on the disconnected problem, while our proposed method can still accurately approximate the Pareto front. The above results show that our proposed method performs well on different kinds of problems and outperforms most of the baselines.

\subsubsection{Image Classification \label{sec33}}
 As shown in Table \ref{visionft}, our proposed method not only has comparable running time and model parameters but also achieves better HV values than all baseline methods. The Pareto fronts obtained by all methods are shown in Fig. \ref{fig:comp}. These figures show that our proposed methods outperform baselines in accurately approximating the Pareto front on Multi-MNIST and Multi-Fash.+MNIST. For Multi-Fashion, most preference vectors are sampled in the middle of the Pareto front because a large number of observation points are clustered in that region. Therefore, most solutions are in the middle of the Pareto front.

\begin{figure}[t]
    \centering
    \includegraphics[width=0.45\textwidth]{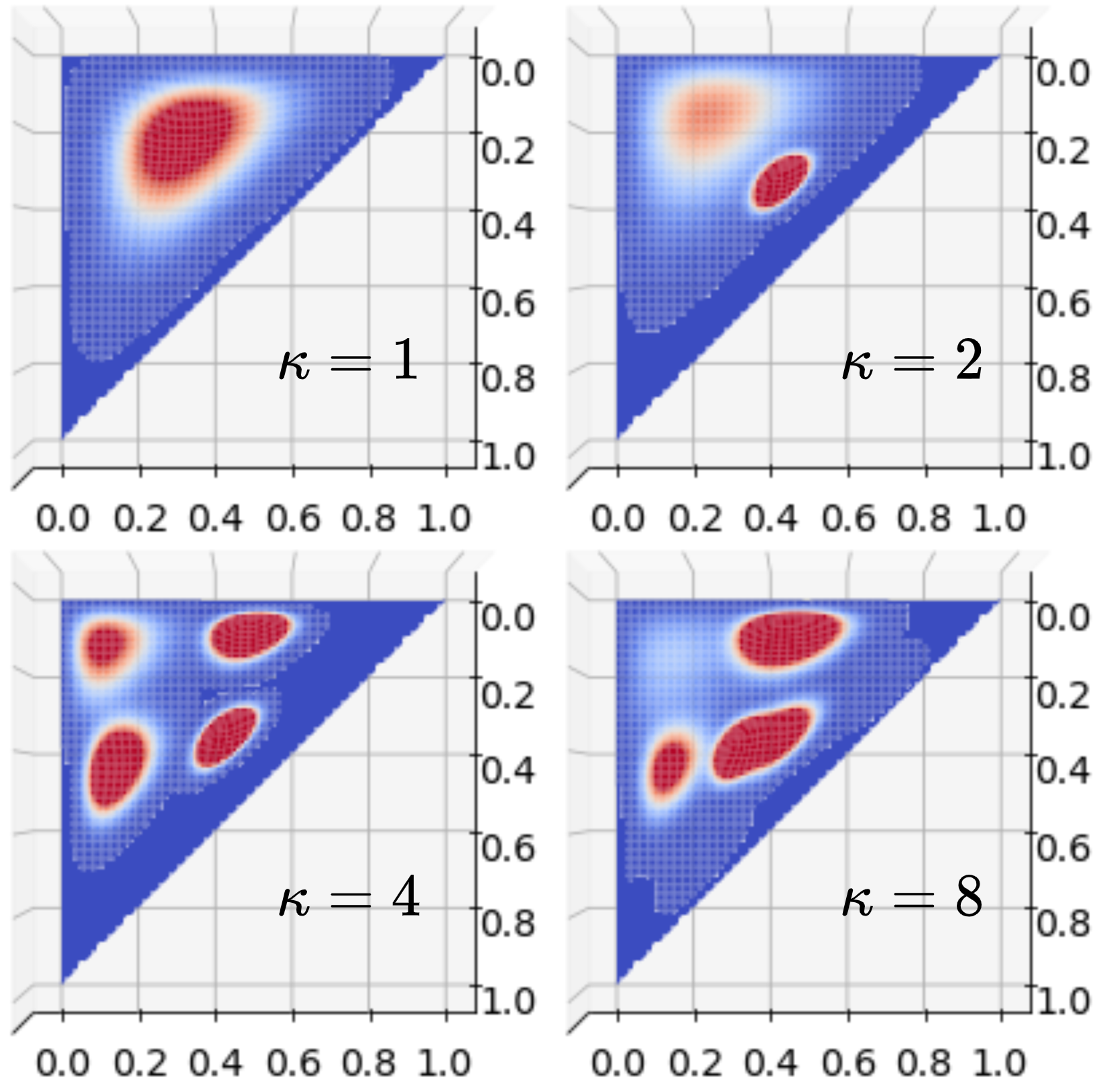}
    \caption{The mixture of Dirichlet distributionss for different $\kappa$ on DTLZ7. The darker the color, the higher the probability of sampling.}
    \label{fig:k}
\end{figure}

\begin{figure}[t]
    \centering
    \includegraphics[width=0.4\textwidth]{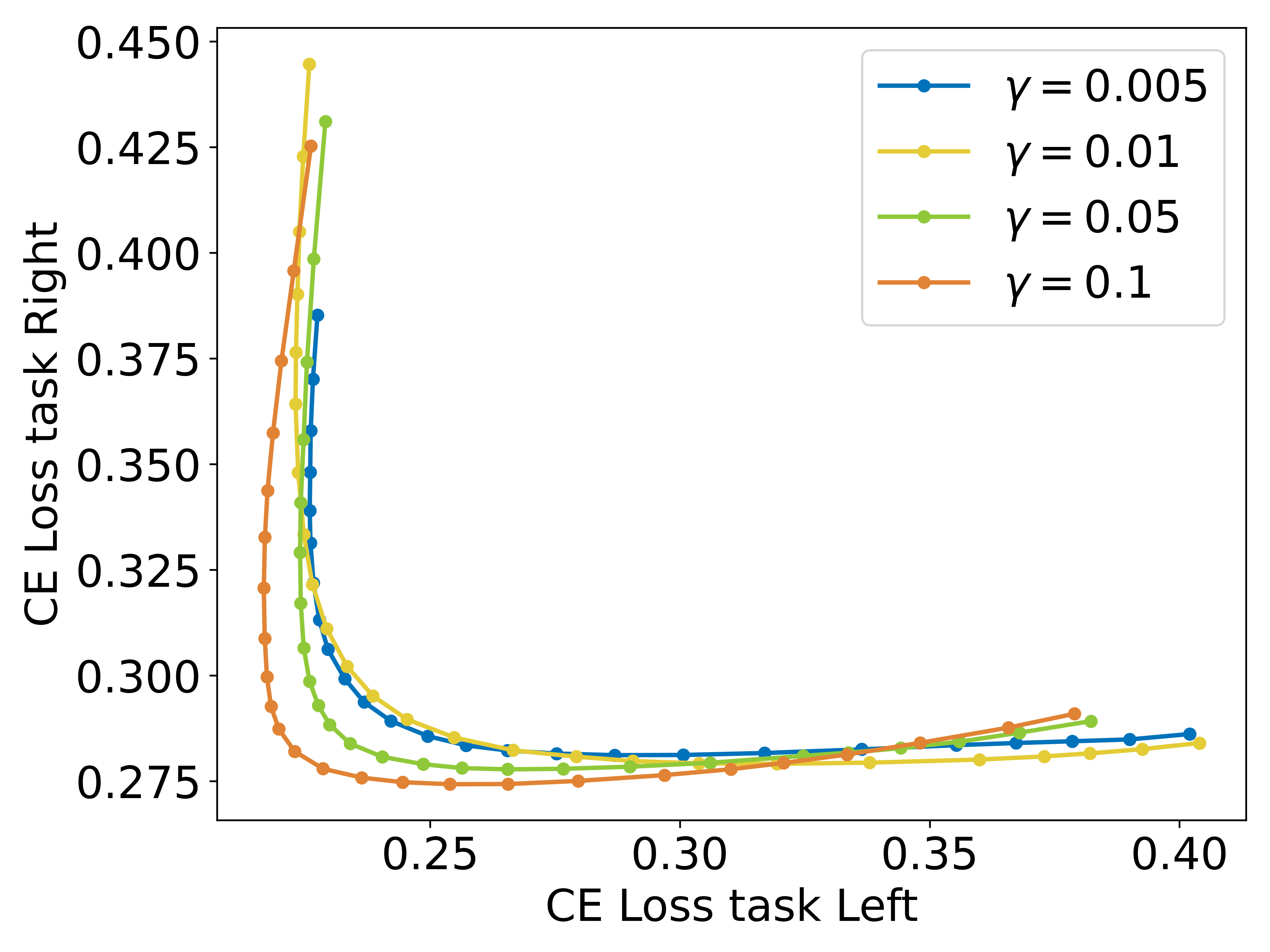}
    \caption{The impact of hyperpareter $\gamma$.}
    \label{fig:sen}
\end{figure}

\subsubsection{Ablation Study \label{sec34}}
To investigate the impact of different $\gamma$ settings on the performance of DDPS-MCMC, we examined DDPS-MCMC with various values of  $\gamma$ on Multi-MNIST. As shown in Fig. \ref{fig:sen}, when $\gamma$ has a large value, there are more dominant solutions in the estimated Pareto fronts at both ends. This is because most observation points tend to cluster in the middle, which increases the probability of sampling preference vectors from the middle of the Pareto front. Larger $\gamma$ can improve the convergence of the obtained Pareto front and correspondingly reduce the diversity of the obtained Pareto front. Therefore, selecting a suitable value of $\gamma$ is helpful for balancing the convergence and diversity of the obtained Pareto front.

To analyze the effect of the number of components $\kappa$ in a mixture of Dirichlet distributions, we examined different values $\kappa$ on DTLZ7. As shown in Fig. \ref{fig:k}, when $\kappa$=1, the sampling density is concentrated in one area. When $\kappa$=2, the sampling density is concentrated in two areas. When $\kappa$=4, the high-density area formed basically corresponds to the position of the Pareto front of DTLZ7. When $\kappa$=8, it should be expected to generate four high-density areas, but three are generated. This may be because the mixture of Dirichlet distribution with $\kappa$=8 has too many parameters, but it still covers the area of the Pareto front.

\section{Related Work \label{S5}}
Multi-Task Learning (MTL) uses a single model to learn multiple tasks, and the parameters of model are shared across tasks \cite{ruder2017overview,zhang2021survey}. In this way, In some cases, MTL-based models outperform single-task models. \cite{standley2020tasks}. Recently, MTL methods have been proposed for adaptive load balancing of competing losses \cite{chen2018gradnorm}, task uncertainty \cite{kendall2018multi}, the rate of change in losses \cite{liu2019end}, and learning non-linear loss combinations by implicit differentiation \cite{navon2020auxiliary}. Nevertheless, these methods only aim to balance all tasks to achieve a compromise solution and are not suitable for modeling task trade-offs.

Multi-Objective Optimization (MOO) aims to find a set of Pareto-optimal solutions based on the trade-offs for different objectives \cite{ehrgott2005multicriteria}. MOO has a wide range of applications \cite{van2014multi,pirotta2016inverse,parisi2016multi,parisi2014policy}. An approach \cite{desideri2012multiple} has been proposed to utilize a gradient-based method for multi-objective optimization. Based on this gradient-based method, many algorithms have been proposed to train multiple neural networks to generate multiple Pareto-optimal solutions at inference time by using multiple preference vectors \cite{mahapatra2020multi,lin2019pareto}. Unfortunately, these method requires excessive training time and are unable to generate a continuous Pareto front. To address this issue, one idea is to utilize an additional hypernetwork to receive preference vectors and generate the parameters of the target network, such as PHN-LS and PHN-EPO \cite{navon2020learning}. The other idea is to consider preference vector information in the feature space, including COSMOS \cite{ruchte2021scalable} and GMOOAR \cite{chen2022multi}.

\section{Conclusion \& Future Work\label{S6}}
In this paper, we proposed a data-driven preference vector sampling method, namely DDPS-MCMC. DDPS-MCMC includes a mixture of Dirichlet distributions with adaptive parameters. DDPSL-MCMC can be applied to disconnected Pareto fronts or different shapes of Pareto fronts. In addition, preference vector sampling is made efficient based on \textit{Non-Dominated Sorting} and \textit{Crowding Distance}. We examined the effectiveness of the proposed method on five synthetic problems and three multi-task image classification problems. Experimental results showed that our method outperformed most of the baselines on most multi-objective optimization problems. However, obtaining high-quality preference vectors for updating the sampling distribution on the more complex Pareto front is still a challenge. Our method is currently not suitable for constrained multi-objective optimization problems. In the future, a potential research direction is to design a suitable sampling strategy for preference vectors to tackle multi-objective optimization problems with more complex Pareto fronts or constraints. 

\section*{Acknowledgment}
This work was supported by National Natural Science Foundation of China (Grant No. 62106099, 62250710163, 62376115), Guangdong Basic and Applied Basic Research Foundation (Grant No. 2023A1515011380), Science and Technology Program of Guangzhou (202201010377), Guangdong Provincial Key Laboratory (Grant No. 2020B121201001).

\bibliographystyle{IEEEtran}
\bibliography{reff}

\vfill

\end{document}